\UseRawInputEncoding

\documentclass[utf8]{FrontiersinVancouver} 

\usepackage{url,hyperref,lineno,microtype,subcaption}
\usepackage[onehalfspacing]{setspace}

\usepackage{appendix}
\usepackage{booktabs}
\usepackage{multirow}
\usepackage{multicol}

\def\firstAuthorLast{Zihao Zhao\,$^{1,2}$} 
\def\Authors{Zihao Zhao\,$^{1,2}$, Yanhong Wang\,$^{1,2}$ , Qiaosha Zou\,$^{1}$ , Tie Xu\,$^{3}$ , Fangbo Tao\,$^{3}$ , Jiansong Zhang\,$^{1}$ , Xiaoan Wang\,$^{4}$ , C.-J. Richard Shi\,$^{1}$ , Junwen Luo\,$^{2,4,*}$ and Yuan Xie\,$^{2}$}
\def\Address{$^{1}$Fudan University, Shanghai, China \\
$^{2}$Alibaba DAMO Academy, Shanghai, China \\
$^{3}$Alibaba Group, Hangzhou, China \\
$^{4}$BrainUp Research Lab, Shanghai, China}
\def\corrAuthor{Junwen Luo}

\def\corrEmail{luojunwen@naolubrain.com}

\begin{document}
\onecolumn
\firstpage{1}

\title[The Spike Gating Flow]{The Spike Gating Flow: A Hierarchical Structure Based Spiking Neural Network for Online Gesture Recognition} 

\author[\firstAuthorLast ]{\Authors} 
\address{} 
\correspondance{} 

\extraAuth{}

\maketitle

\begin{abstract}
Action recognition is an exciting research avenue for artificial intelligence since it may be a game changer in the emerging industrial fields such as robotic visions and automobiles. However, current deep learning faces major challenges for such applications because of the huge computational cost and the inefficient learning. Hence, we develop a novel brain-inspired Spiking Neural Network (SNN) based system titled Spiking Gating Flow (SGF) for online action learning. The developed system consists of multiple SGF units which assembled in a hierarchical manner. A single SGF unit involves three layers: a feature extraction layer, an event-driven layer and a histogram-based training layer. To demonstrate the developed system capabilities, we employ a standard Dynamic Vision Sensor (DVS) gesture classification as a benchmark. The results indicate that we can achieve 87.5\% accuracy which is comparable with Deep Learning (DL), but at smaller training/inference data number ratio 1.5:1. And only a single training epoch is required during the learning process. Meanwhile, to the best of our knowledge, this is the highest accuracy among the non-backpropagation algorithm based SNNs. At last, we conclude the few-shot learning paradigm of the developed network: 1) a hierarchical structure-based network design involves human prior knowledge; 2) SNNs for content based global dynamic feature detection. 

\end{abstract}

\section{Introduction}
\label{sec:intro}

Deep Learning (DL) nowadays exerts a substantial impact on a wide range of computer vision tasks such as face recognition \cite{DBLP:conf/iccvw/HuYYKCLH15} and image classifications \cite{DBLP:conf/nips/KrizhevskySH12}. But it is still facing major challenges when processing information with high dimensional spatiotemporal dynamics such as video action recognition. This is because of: 1) huge computational cost: the deep neural networks have to capture dynamic information across another timing dimensions, which requires significant computational resources for the training stage \cite{DBLP:conf/cvpr/HeZRS16}; 2) inefficient learning: events contain significant global dynamic features that are seldomly captured by the DL, but these can be easily recognized by biological systems \cite{Purves4750}. One promising technology of sparsity \cite{DBLP:conf/cvpr/LiuWFTP15, DBLP:conf/nips/WenWWCL16, DBLP:conf/glvlsi/LiuZWZZS21} can relieve the first issue of the intensive computing to some extend, but the training cost is still enormous. Recently, Few-Shot Learning (FSL) is proposed to tackle this problem. With prior knowledge, FSL can quickly generalize to new tasks with only a few labeled training samples \cite{fsl_survey, HGNNFSL, RNFSL}.

Fortunately, Spiking Neural Networks (SNNs) is an alternative candidate to perform spatiotemporal related tasks \cite{DBLP:journals/nn/LoboSBK20} with few-shot learning capability. By taking the nature characters of computing with time and an event-driven manner, real-world event information will be encoded into spike trains as inputs along with timing frames. With a brain-inspired hierarchical network architecture processing, the output spike patterns are interpreted as inference results via neural decoding methods (e.g. spike timing coding, rank coding, spike count coding). Therefore, this could be an efficient technique for such applications, and another potential path towards the next generation AI \cite{DBLP:series/sci/FurberT08}.

\begin{figure}[t]
	\centering
	\includegraphics[width=0.6\columnwidth]{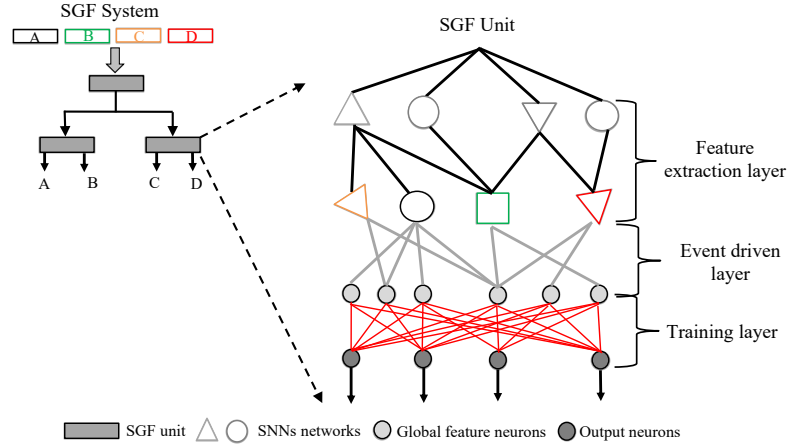} 
	\caption{The Spike gating flow system concept. It consists of multiple SGF units. Each SGF unit involves with three layers: 1) a feature extraction layer, 2) an event-driven layer, 3) a training layer. On the top left there is an example of a four-event classification SGF system: A, B, C, D are four event types.}
	\label{Fig-sgf_system}
\end{figure}

However, employing SNNs for action recognition remains challenging since it lacks an efficient learning algorithm. Recently SNN based learning systems can be classified into three levels: a micro-level, a middle-level and a macro-level system. A micro-level based systems emphasis on utilizing low-level spiking neuron computing characters such as a temporal process and an integration-and-fire manner \cite{10.3389/fnins.2018.00331, DBLP:conf/cvpr/AmirTBMMNNAGMKD17, 10.3389/fnins.2016.00508, DBLP:conf/nips/ZhangL19, STDP}. For instances, \cite{10.3389/fnins.2018.00331} proposes an SNN based spatiotemporal Back-Propagation (BP) for a dynamic N-MINST event classifications, the developed algorithm successfully combines spatial and time domain kernels and achieves inference accuracy of 98.78\%. Also, \cite{DBLP:conf/cvpr/AmirTBMMNNAGMKD17} illustrates a Convolution Neural Network (CNN) based SNN for gesture classification. By employing an event-driven sensor and a TrueNorth neuromorphic chip, the system shows 178.8mW power consumption and 96.49\% accuracy. Meanwhile, A reservoir layer-based SNN utilizes Spike-timing dependent plasticity rule to update weights. Such a novel network can achieve 95\% 
top-3 accuracy on IBM DVS gesture task \cite{DBLP:conf/cvpr/AmirTBMMNNAGMKD17}. However, the SNN higher-level computing entities such as attractor dynamics are missed in the system, which results in inefficient learning.

A middle-level system indicates SNNs apply global dynamic behaviors on the learning process \cite{10.1162/0899766053630332, 10.3389/fninf.2013.00048, DBLP:conf/nips/VoelkerKE19, DBLP:journals/corr/abs-2110-02402, DBLP:journals/corr/abs-2001-10159, SUSSILLO2009544}. A FORCE learning method \cite{SUSSILLO2009544} is able to convert network chaos patterns into a desired one by modifying synaptic weight. Also, a Neural Engine Framework (NEF) develops a method to build dynamic systems based on spiking neurons \cite{10.3389/fninf.2013.00048}. Such an approach leverages neural non-linearity and weighted synaptic 
filter as computational resources. Compared to the first type SNNs, a middle-level based SNNs emphasis on global network dynamic rather than individual spiking neurons computing characters. Therefore, such systems demonstrate much better learning behaviors regarding scalability \cite{DBLP:conf/nips/VoelkerKE19} and model sizes \cite{DBLP:journals/corr/abs-2110-02402} at some particular scenarios. 

A macro-level system includes both micro-level and middle-level system's advantages \cite{SUSSILLO2009544, DBLP:journals/corr/abs-1906-07067}. It combines detailed spiking neuron characters and network dynamics together to form a unique learning system. For instance, an olfactory SNN is largely based on the mammalian bulb network architecture, and with a line attractor based neural plasticity rule for online learning odorants \cite{DBLP:journals/corr/abs-1906-07067}. The developed system shows great one-shot learning behavior compared to the DL. Meanwhile, \cite{wu2022brain} proposed a spike-based hybrid plasticity model for solving few-shot learning, continual learning, and fault-tolerance learning problems, it combines both local plasticity and global supervise information for multi-task learning.

In this work we develop a novel macro-level system titled Spike Gating Flow (SGF) for action recognition as shown in Fig. \ref{Fig-sgf_system}. The system consists of multiple SGF units that connect in a hierarchical manner. An SGF unit consists of three layers: 1) a feature extraction layer for global dynamic feature detection; 2) an event-driven layer for generating event global feature vectors; 3) a supervise-based histogram training layer for online learning (redlines in Fig. \ref{Fig-sgf_system}). By employing a Dynamic Vision Sensor (DVS) \cite{DBLP:journals/jssc/PoschMW11} based gesture dataset \cite{DBLP:conf/cvpr/AmirTBMMNNAGMKD17} as a benchmark, the results demonstrate that the developed SGF has great learning performance: 1) the system can approximately achieve the same level accuracy of 87.5\% as the DL but at a training/inference sample ratio 1.5:1 condition. More importantly, only one epoch is required during the training; 2) to our best of knowledge, this is the highest accuracy among the non-BP based SNNs; 3) the system consumes only 9mW and 99KB memory resources on an FPGA board at the inference stage. In summary, the contributions are as follows:

\begin{itemize}
\item Algorithm aspect: We developed an efficient few-shot learning system for gesture recognition, which behaves like the biological intelligence: few-shot learning, energy efficient and explainable. 
\item Application aspect: the SGF based hardware shows reasonable memory size (99KB) and power consumption (9mW), which is suitable for the edge/end-device scenarios.
\item Learning theory aspect: We conclude  one few-shot learning paradigm: 1) a hierarchical structure-based network design involves with human prior knowledge; 2) SNNs for global dynamic feature detection. 
\end{itemize}

\section{The Spike Gating Flow}
\label{sec:sgf}
The Spike Gating Flow (SGF) is a new learning theory to achieve online few-shot training entities, which is inspired from the Neural Engineering Framework (NEF) \cite{NEF} and brain assemble theories \cite{Papadimitriou14464}. In brief, the few-shot learning capabilities rely on prior knowledge embedded in the hierarchical architecture and global feature computing. While the online computing benefits from using dynamic
spike pattern to encode both data and control flow. Therefore, network different level nodes are served as gates to pass or stop input data information, and spikes are served as gate control signals. We have concluded the key principles of SGF as below:
\begin{itemize}
    \item \textbf{Global feature representations}: Network representations are defined by the combination of different SNNs global movement features rather than pixel local features.
    \item \textbf{Tailor designed hierarchical network structure}: A hierarchical structure-based network for conditional data-path execution. Depending on inputs, SGF unit spike patterns are served as gates command to manipulate data-paths.
    \item \textbf{Histogram based training algorithms}: A global feature-based histogram training adjusts output layer weights based on history information.
\end{itemize}

\begin{figure}[t]
	\centering
	\includegraphics[width=0.6\columnwidth]{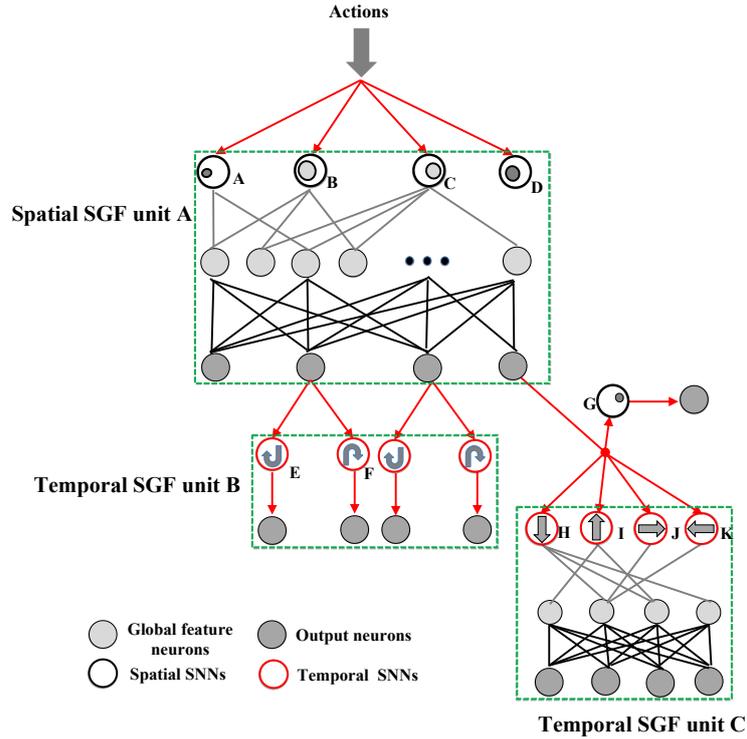} 
	\caption{The SGF network architecture. It mainly consists of three SGF units: a spatial SGF unit A and temporal SGF units B and C. A spatial SGF unit A has four SNNs with feature ID index A-D (A: intensive activities at constrained left areas; B: mild activities at plateau left areas; C: mild activities at plateau right areas D: intensive activities at constrained right areas). A temporal SGF unit B has two SNNs with feature ID index E-F (F: clockwise movement; G: clockwise counter movement). A temporal unit C has four SNNs with feature ID index H-K (H: top-down; I: bottom-up; J: left-right; D: right-left). Also, the developed network has 10 output neurons corresponding to 10 action types.}
	\label{Fig-sgf_network}
\end{figure}

Based on such principles, we design three SGF units and carefully connect them into a two-level network, particularly for online gesture recognition. Such an architecture can be considered as a pre-designed learning rule, and each SGF unit is designed based on the cell assemblies theories \cite{mullerENEURO.0533-19.2020}, and responses for a unique content based global dynamic features.  This aims that the system can manipulate units into a hierarchical level to maximize learning efficiency. As shown at Fig. \ref{Fig-sgf_network}, the top area of the developed network is a spatial SGF unit A, and the bottom areas are temporal SGF units B and C. Typically, a SGF unit consists of three layers: a feature extraction layer, an event-driven layer and a histogram-based training layer. Also, there can be some structure variations of SGF units. For instance, an SGF unit B has a feature extraction layer only. 

For an SGF unit computing process, at first a feature extraction layer is to detect events global dynamic features. Each SGF unit has several corresponding spatial SNNs and temporal SNNs, which targets on detecting different global features (features with index A-I are shown at Fig. \ref{Fig-sgf_network}). An SGF unit A has five spatial SNN networks, which responses for detecting spatial features as below: A) intensive activities at a left constrained area; B) mild activities at a left large area; C) mild activities at a right large area; D) intensive activities at right constrained areas; G) intensive activities at a specific constrained area. And an SGF unit B has two temporal SNN networks: E) clockwise movements and F) clockwise counter movements. Particularly, a human prior knowledge of event sequences is introduced here for designing E and F. An SGF unit C has four temporal SNNs: H) up-down movements; I) bottom-up movements; J) left-right movements and K) right-left movements. The detailed computing mechanisms of SNNs are described at the Section 3 of SNN design.

Next there is an event-driven layer that connects SNNs outputs to the global feature neurons. This layer responses for generating event feature vectors for the next layer training. Typically, an event class will have several feature vectors types due to the spatiotemporal variations. A feature vector can be defined as a combination of active SNNs’ feature index, which are represented by connecting active SNNs to one global neuron. Therefore, for each action type, global feature neuron number is equal to the action type feature vector type number.

At last, an SGF unit has a fully connected histogram-based training layer, in which each output neuron connects to its all global feature neurons. After each training trail, feature vector histogram numbers will be updated and converted into corresponding weights. And the conversion is a normalization process. This result of the higher histogram number is, the bigger weights are. At an inference stage, a test sample generated feature vector will be sent into all output neurons for calculating final scores, which follows the equation as below: 

\begin{equation}
\label{equ:sgf}
S_{m}=\sum_{j} \frac{\left(T_{j}^{m} * V\right)}{L_v} \times w_{m}^{j}
\end{equation}

Where $S_m$ is a testing sample score at $m^{th}$ output neuron; $w_m$ is the $j^{th}$ feature vector weights of the $m^{th}$ output neuron; $T_j^m$ is the $j^{th}$ feature vector of the $m^{th}$ output neuron; and $V$ is the feature vector of the testing samples. The symbol $*$ is a bit-wise NOR operation, and $L_v$ is the bit length of the feature vector. The key advances of such a learning algorithm are that each data sample only requires one training time and tiny computational resources for updating weights, which enables rapid online learning behaviors.

\begin{figure*}[t]
	\centering
	\includegraphics[width=1.0\columnwidth]{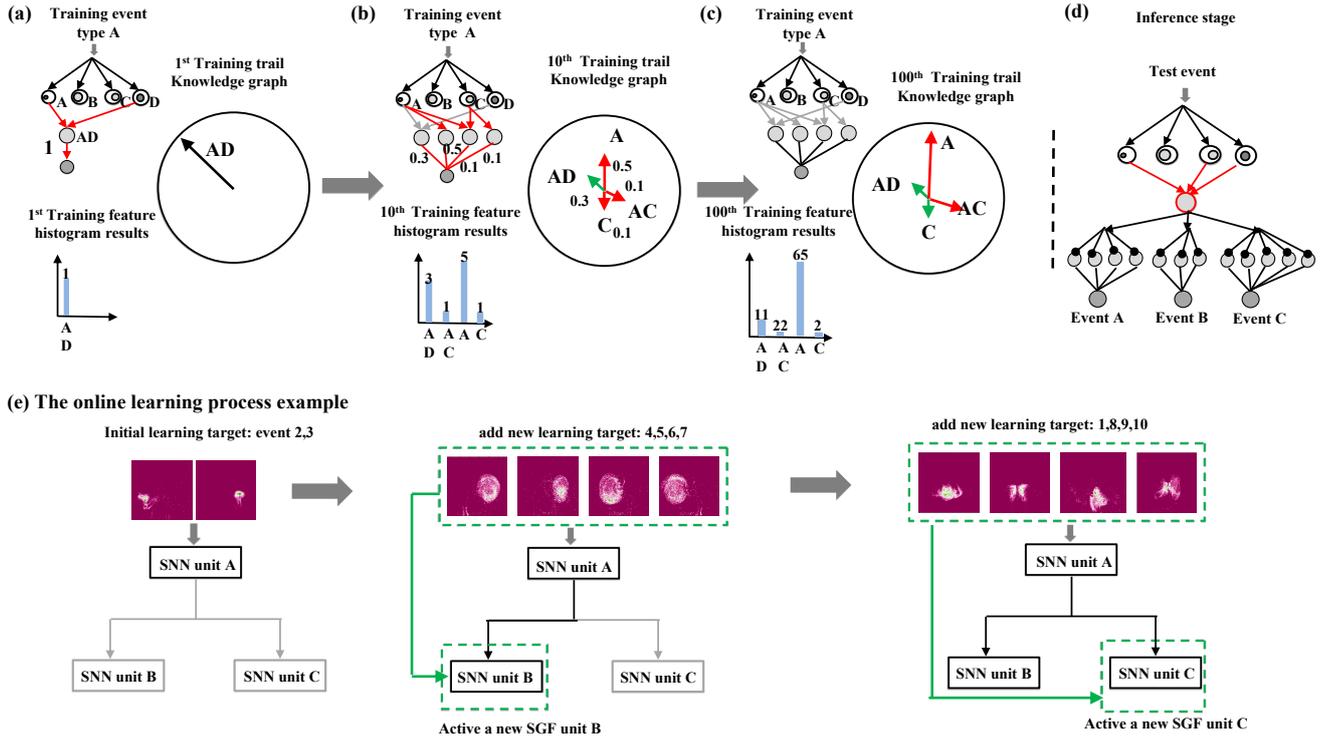} 
	\caption{(a)-(d) The histogram-based training example of an event type A; (e) The system online learning process example.}
	\label{Fig-learning}
\end{figure*}

A detailed example is illustrated at Fig.\ref{Fig-learning}. SNNs with feature index $A$ and $D$ are active at the first training trail, which forms a feature vector$[A-D]$. Hence a corresponding global feature neuron is generated that connects to SNNs with feature index $A$ and $D$ (connected with red lines). And a feature vector histogram is also displayed on the Fig. \ref{Fig-learning}(a) left. After that, the feature vector$[A-D]$ histogram values will be converted into event type $A$ output neuron weights. It is clearly seen that the weight is one since there is only one feature vector type (Fig. \ref{Fig-learning}(a)). Meanwhile a knowledge graph of event type $A$ is produced for quantitative analysis feature vector distributions (Fig. \ref{Fig-learning}(a) right). At a $10th$ training trail, there are  three more feature vectors generated $[A-C, A, C]$ (Fig. \ref{Fig-learning}(b) left red lines). This indicates that there are in total four types of feature vector in the event type $A$. Identically, corresponding feature vectors histogram numbers $[3,1,5,1]$ will be transformed into event $A$ neuron outputs weights via a training layer. The feature vectors distribution is also updated in the knowledge graph: a vector with green lines indicates histogram values are decreased, while a vector with red lines indicates histogram values are raised. At the end of a 100th training trail, there is no new feature vector appeared, which results of the same global feature neuron number as the 10th training trail. The event $A$ output neuron weights are updated based on the current histogram numbers as a final result. Similarly, the other event types B and C follow the same training procedures. 

At the inference stage, a test sample is given into the trained network, which will generate a corresponding test feature vector. And it will go through all the output neurons to calculate the final scores. As shown at Fig. \ref{Fig-learning}(d), there is a trained network which contains three output neurons, whose inference classification result is the maximum one among these output neuron scores. 

The online learning process example is shown at Fig. \ref{Fig-learning}(e). At an initial stage, event group A [3: right hand wave; 2: left hand wave] is sent into the network for training. Since event group A contains significant spatial features, only a spatial SGF unit A is active and responsible for generating feature vectors. After finishing learning event group A, event group B [4: right arm clockwise, 5: right arm counter clockwise, 6: left arm clockwise, 7: left arm counter clockwise] is sent into the network for sequential online learning. Identically, a temporal SGF unit B is active for recognizing clockwise/counter clockwise movements. At last, event group C [1: hand clap, 2: left hand wave, 8: arm rolls, 9: air drum, 10: air guitar] is sent into network that contains complex combinations of vertical and horizontal movements. The rest of SGF unit C is active for learning such features. As it can be seen, the final network architecture varies depending on the learning targets.

\section{SNNs Design}
\label{sec:snns}

 We have developed three SNN types that are SpatioTemporal (ST) cores, spatial SNNs and temporal SNNs. An ST core aims to reduce the DVS camera background noise and enhance key spatiotemporal information. Spatial SNNs response for capturing event's spatial features. And temporal SNNs expert on distinguishing object movement directions. By combining these SNNs, a system can have rich representations of various spatiotemporal features for action classifications. 

\subsection{Spatiotemporal Core}

Since a DVS camera has a unique output format, a spatiotemporal core is designed for DVS output preprocessing: 1) to reduce the DVS output noise; and 2) to enhance the key spatiotemporal information. An ST core involves with two-stage computations: a spatial and temporal processing. The mathematical model is as below:

\begin{equation}
\label{stcore}
S T_{m}^{t}=\left[\int_{t-\Delta S T_{t}}^{t}\left[\sum_{i}^{i+\Delta S T_{s}} d_{m}^{t}\right]^{\theta_{s}} d t\right]^{\theta_{t}}
\end{equation}

Where ${ST}_m^t$ is the outputs of the $m^{th}$  ST core at frame $t$ period;  $d_m^t$  is the outputs of the $m^{th}$  DVS sensor pixel at frame $t$,  which equals to -1 or +1;  $\Delta{ST}_s$ is a ST core spatial detection range. The function ${[S]}^{\theta_s}$ equals $1$ if S over spatial thresholds $\theta_s$. Regarding the temporal computations, $\Delta{ST}_t$ is an integration window and $\theta_t$ is a temporal threshold. The function $[T]^{\theta_t}$  equals $1$ if $T$ is over spatial thresholds $\theta_t$. As a result of this, by adjusting above four parameters $[\Delta{ST}_s$  ,$\theta_s$,$\Delta{ST}_t,\theta_t]$, we can configure ST core filtering behaviors properly. In general, each SNN will require a ST core for feature extraction. More details of preprocessing DVS gesture dataset are illustrated in the Results section 4.3.   

\subsection{Spatial SNNs}

\begin{figure}[t]
	\centering
	\includegraphics[width=0.55\columnwidth]{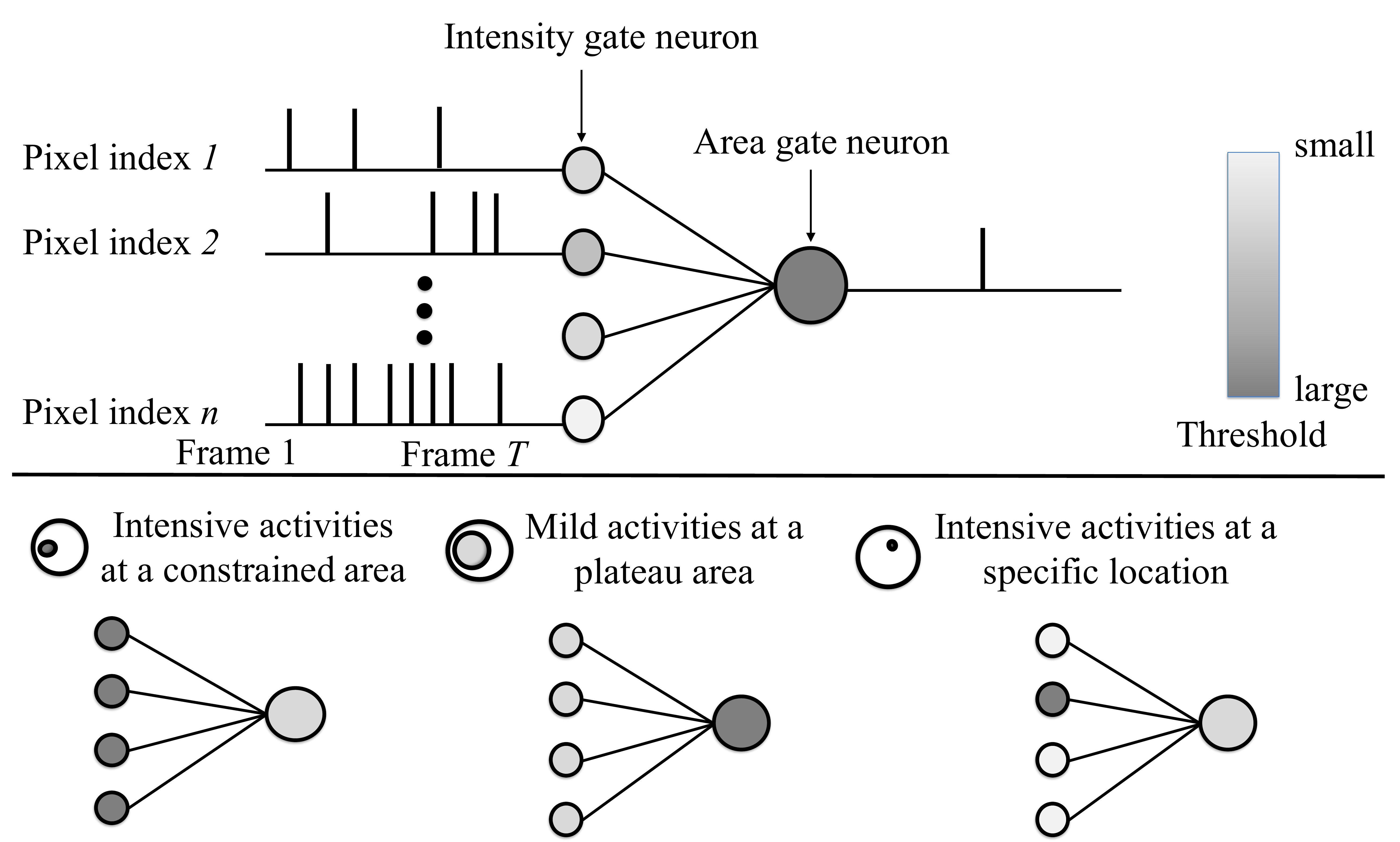} 
	\caption{(a) The spatial SNN computing mechanism. The event consists of T frames in total. (b) Examples of three spatial SNNs focus on different global feature detection.}
	\label{Fig-spatial_snn}
\end{figure}
A spatial SNN is designed for extracting spatial features based on ST core outputs.  The computing mechanism is quite similar with an ST core. However, the major differences rely on 1) the outputs of a spatial SNN are a feature vector; 2) A spatial SNN does not require temporal information, and it accumulates all the frames [0, T] together first and performs spatial computing. The spatial SNN equations are as below:

\begin{equation}
\label{spatial_snn}
S P_{m}=\left[\sum_{i}^{i+\Delta S P_{s}}\left[\int_{t=0}^{t=T} S T_{m}^{t} d S T\right]^{\theta_{i}}\right]^{\theta_{a}}
\end{equation}

Where $T$ is the total frame number of an event. $\Delta{SP}_s$ is the  detection size and ${SP}_m$ is the $m_{th}$ spatial SNN outputs. And the outputs can be a single bit or multiple bits. $\theta_i$ is an intensity gate neuron threshold, $\theta_a$ is an area gate neuron threshold. As Fig. \ref{Fig-spatial_snn}(a) depicts, an event consists of T frames is processed by gate neurons first. After that, gate neuron spikes are employed as inputs for area gate neurons.

By adjusting two thresholds’ values, three different spatial SNNs can be reconstructed at Fig. \ref{Fig-spatial_snn}(b): 1) spatial SNNs with feature index $[A, D]$: intensive activities as a constrained area $[\theta_1>\theta_2]$; 2) spatial SNNs with feature index $[B, C]$: mild activities at a large area $[\theta_1<\theta_2]$;  and 3) spatial SNNs with feature index $[G]$:  intensive activities at a specific location.

\subsection{Temporal SNNs}

Temporal SNNs are designed for detecting movement directions of an event (e.g., up-down, left-right). The key principle is to encode object's temporal location into temporal neuron spiking patterns. The equation is shown below: 

\begin{figure}[t]
	\centering
	\includegraphics[width=0.55\columnwidth]{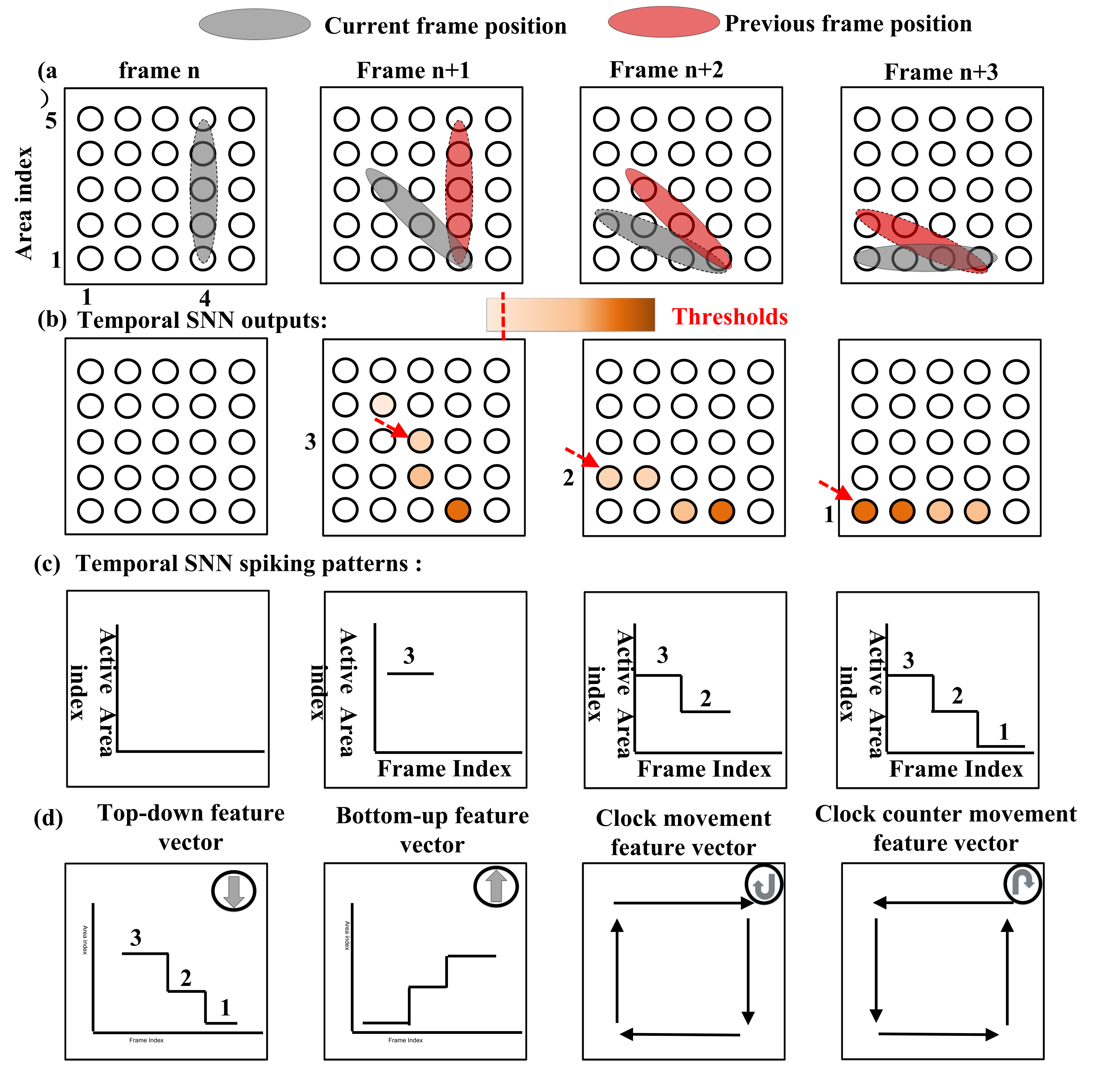} 
	\caption{The temporal SNN with feature index H top-down computing mechanism. Also, reference patterns of temporal SNNs with feature index E, F and I are shown at bottom as well.}
	\label{Fig-temporal_snn}
\end{figure}

\begin{equation}
\label{temporal_snn}
TE_{m}^{t}=\left[\sum_{i}^{n^{t-\Delta TE_{t}}}\left[l_{m}^{t}-l_{i}^{t-\Delta TE_{t}}\right]^{\theta_{l}}\right]^{\theta_{te}}
\end{equation}

Where ${TE}_m^t$ is the outputs of the $m^{th}$  temporal neuron at frame $t$; $l_i^t$ is the location of the $i^{th}$  temporal active neuron at frame $t$, the location can be either vertical or horizontal information depends on temporal SNN types. $\Delta{TE}_t$ is the comparison frame window; $\theta_l$ is the location index threshold,  $n^{t-\Delta{TE}_t}$ is the active neuron number at frame $t-\Delta{TE}_t$. $\theta_{te}$ is the temporal neuron spiking threshold. For each frame, the object location is calculated by selecting the maximum location index of fired neurons. Therefore, objects' temporal movement patterns can be obtained by a combination of generated object locations at each frame, which serve as event temporal feature vectors. 

For instance, Fig. \ref{Fig-temporal_snn} shows a temporal SNN with feature index $H$ of detecting top-down movements. At a frame $n$, an object is vertically located at areas from $[4,1]$ to $[4,5]$. Since this is the first frame, a temporal SNN does not generate activities because there is no comparison reference. At a frame $n+1$, an object is moving down to a new location as the gray color indicates.  Each active neuron (neuron who receives non-zero ST core outputs) will compare its location to all the active neurons at the previous compared frame. If the current neuron location is lower than the previous frame neuron location, the current neuron will receive an input value from the compared neuron. In this top-down case, the location index is defined as the vertical information (Y-axis). The temporal SNN outputs are shown at Fig. \ref{Fig-temporal_snn}(b), neurons with dark colors indicate large input values, while neurons with light color indicate limited input values. Neurons then generate a spike when values are above a threshold $\theta_{te}$. After calculating all the frames, the object's temporal movements are represented by the generated temporal feature vectors. A temporal SNN with feature index $H$ will be active if an object movement temporal feature vector consists with its reference feature vector which is shown at Fig. \ref{Fig-temporal_snn}(d) bottom (top-down feature vector). Identically, temporal SNNs with feature index $[I,J,K]$ follow the same computing mechanisms. Corresponding reference feature vectors are also shown at \ref{Fig-temporal_snn}(d) bottom. Reference feature vectors which can be pre-defined or learned depends on the task. Particularly,  SNNs with feature index $[E,F]$ clockwise and counter clockwise require temporal feature vector timing information that is based on a human prior knowledge. The clockwise counter event temporal pattern sequence is defined as [top-down, left-right, bottom-up, right-left], and the clockwise event temporal pattern sequence is defined as [top-down, right-left, bottom-up, left-right].  

\section{Results}
\label{sec:results}

A DVS gesture dataset \cite{DBLP:conf/cvpr/AmirTBMMNNAGMKD17} (10 different gesture actions) is employed to verify the system performance, since such novel DVS based datasets \cite{9138762, 10.3389/fnins.2016.00405} may exert a significant impact on emerging applications such as automobile and robotics. The training and inference dataset ratio is 1.5:1. The event-driven sensor data preprocessing is followed a standard method \cite{DBLP:conf/cvpr/RebecqRKS19}: an event consists of 50-80 frames, and each frame contains 1000 spikes. 

\subsection{Network architecture}

\begin{table}[!t]
    \footnotesize
    \centering
    \begin{tabular}{cccccc}  
    \toprule   
    \toprule   
    SNN module & \multicolumn{3}{c}{Spatial SNN} & \multicolumn{2}{c}{Temporal SNN} \\
    \cmidrule(r){2-4} \cmidrule(r){5-6} 
    index & A/D & B/C & G  & E/F & H/I/J/K \\
    \midrule   
    Number of OPs & 56.0K & 63.0K & 7.0K & 2.1M & 27.2K \\  
    Model size & 304Byte & 342Byte & 2Byte & 240Byte & 16Byte\\ 
    Sub-network number & 16 & 9 & 2 & 80 & 1 \\
    Feature vector length & 16 & 18 & 2 & 160 & 2 \\
    \bottomrule  
    \end{tabular}
    \caption{The developed network architecture information.}
    \label{Tab-snn_module}
\end{table}

 The detailed network architecture information is shown in Tab. \ref{Tab-snn_module}. Each SNN type has different sub-network numbers, and each individual shares identical functionalities but with slightly different parameters. SNNs memory sizes and operation numbers are also displayed in the Tab. \ref{Tab-snn_module}: SNNs with feature index A/D and B/C have the sub-network number 16 and 18, respectively. And the corresponding model sizes are 304 Byte and 342 Byte. Temporal SNNs with feature index E/F have the largest number of OPs: 2.1M. The details of parameter calculation are described at Appendix. A.

\subsection{System Accuracy}

\begin{table*}[!t]\footnotesize
    \centering
    \begin{tabular}{ccccccccccc}  
    \toprule   
    \toprule   
    \multirow{2}{*}{Name} & \multirow{2}{*}{Type} & Learning  & Learning & \multicolumn{4}{c}{Model information}    & \multicolumn{2}{c}{Training cost}  & \multirow{2}{*}{Accuracy} \\
    \cmidrule(r){5-8} \cmidrule(r){9-10} 
    ~&~& method & style & Size  & Diff($\times$) & OPs & Diff($\times$)  & Epoch & T/I ratio      \\
    
    \midrule   
    Reservoir CSNN \cite{DBLP:conf/ijcnn/GeorgeBDMB20} & SNN & STDP & Offline & 3.17MB & $88.7\uparrow$ & - & ~ & - & 3.8:1 & 65.0\%   \\  
    
    Heterogeneity Network \cite{Perez-Nieves2020.12.18.423468} & SNN & SGD & Offline & ~125KB & $3.4\uparrow$  & - & ~ & - & 3.8:1 & 82.1\%\\ 
    
    SCRNN \cite{scrnn} & ANN2SNN & BPTT & Offline & 732.34KB & $20.0\uparrow$ & 81.91M & $9.9\uparrow$ & 100 & 4.1:1 & 96.59\%   \\  
    
    SLAYER \cite{slayer} & ANN2SNN & BP & Offline & 1034.8KB & $28.3\uparrow$  & 79.8M & $9.6\uparrow$ & 739 & 3.8:1 & 93.64\%\\
    
    Converted SNN \cite{ann2snn} & ANN2SNN & BP & Offline & 500KB & $13.7\uparrow$ & 651M & $78.7\uparrow$ & 10 & 3.8:1 & 96.97\%\\

    ConvNet \cite{DBLP:conf/cvpr/AmirTBMMNNAGMKD17} & DNN2SNN & BP & Offline & 16.3MB & $456\uparrow$ & 946.82M & $114\uparrow$ & ~250 & 3.8:1 & 96.5\%\\
    
    PointNet++ \cite{DBLP:conf/cvpr/QiSMG17} & DNN2SNN & BP & Offline & 3.50MB & $98\uparrow$ & 440.0M & $53.2\uparrow$ & ~250 & 3.8:1 & 97.08\%\\
    
    \textbf{This work} & \textbf{SNN} & \textbf{SGF}  & \textbf{Online} & \textbf{36.58KB} & ~ & \textbf{8.27M} & ~ & \textbf{1} & \textbf{1.5:1} & \textbf{87.5\%}   \\
    \bottomrule  
    \multicolumn{10}{l}{"-" indicates the data can not be calculated or not mentioned in the corresponding paper.}
    \end{tabular}
    \caption{The comparison between state-of-the-art methods and the proposed SGF network.}
    \label{Tab-result}
\end{table*}

In Tab. \ref{Tab-result}, we first compare the developed network with two typical SNN-based gesture recognition networks, a STDP based SNN \cite{DBLP:conf/ijcnn/GeorgeBDMB20} and a SGD based SNN \cite{Perez-Nieves2020.12.18.423468}. To the best of our knowledge, our system indicates the highest accuracy of 87.5\% among state-of-the-art non-BP based SNNs. Regarding ANN/DNN converted SNN, the developed network can reach the same level of accuracy as a SLAYER \cite{slayer}, but a slight lower than ConvNet \cite{DBLP:conf/cvpr/AmirTBMMNNAGMKD17} 96.5\%, SCRNN \cite{scrnn} 96.59\%, Converted SNN \cite{ann2snn} 96.97\% and PointNet++ \cite{DBLP:conf/cvpr/QiSMG17} 97.08\%. However, the network model size can be reduced by 456 times compared to the ConvNet \cite{DBLP:conf/cvpr/AmirTBMMNNAGMKD17}, and number of operations can be reduced by 53 times compared to the PointNet++ \cite{DBLP:conf/cvpr/QiSMG17}. 

Last but not the least, the developed SGF only requires 1 training epoch at a condition of training/inference ratio 1.5:1, which DL networks typical require hundreds training epochs at a condition of 3.8:1. This indicates the system training cost is significantly lower than the DL based networks.

\subsection{SNNs Behaviors}

\begin{figure}[th]
    \centering
	\includegraphics[width=0.4\columnwidth]{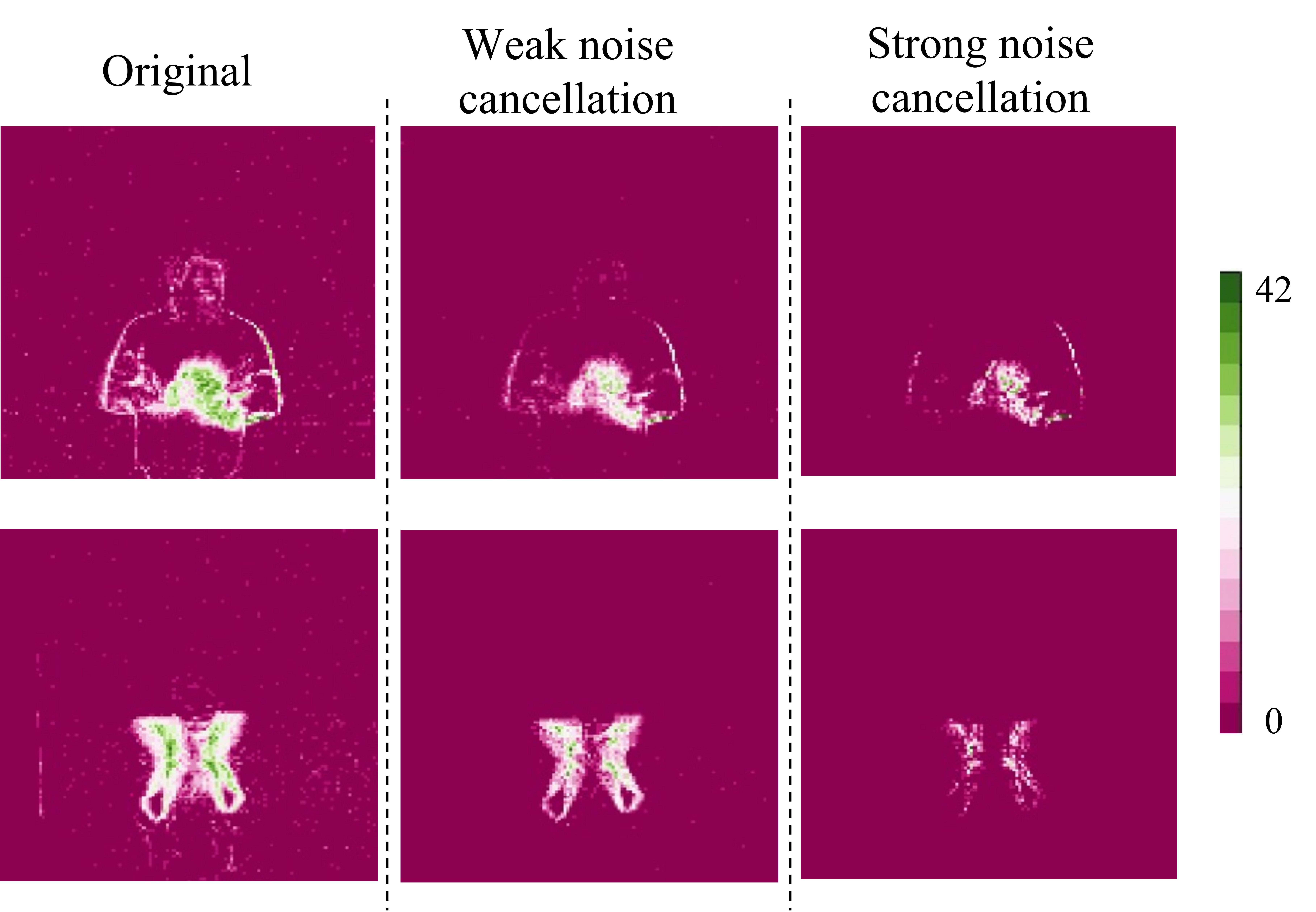} 
    \caption{The ST core weak and strong noise cancellation results.}
    \label{Fig-snns-st_core_demo}
\end{figure}

The ST core noise cancellation performance is shown at Fig. \ref{Fig-snns-st_core_demo}. At left there are original event pictures of a hand clap and an air drum. Event pictures are obtained via a ST core model process. A color bar on the right displayed spike intensities at each pixel. At middle there are results of an ST core weak noise cancellation (parameters are $1, 1, 2, 2$), it is clearly seen that most of sparse noise are disappeared. And at right there are results of an ST core strong noise cancellation (parameters are $1, 1, 6, 5$), only most significant features are kept at this case. 
\begin{figure}[th]
    \centering
	\includegraphics[width=0.3\columnwidth]{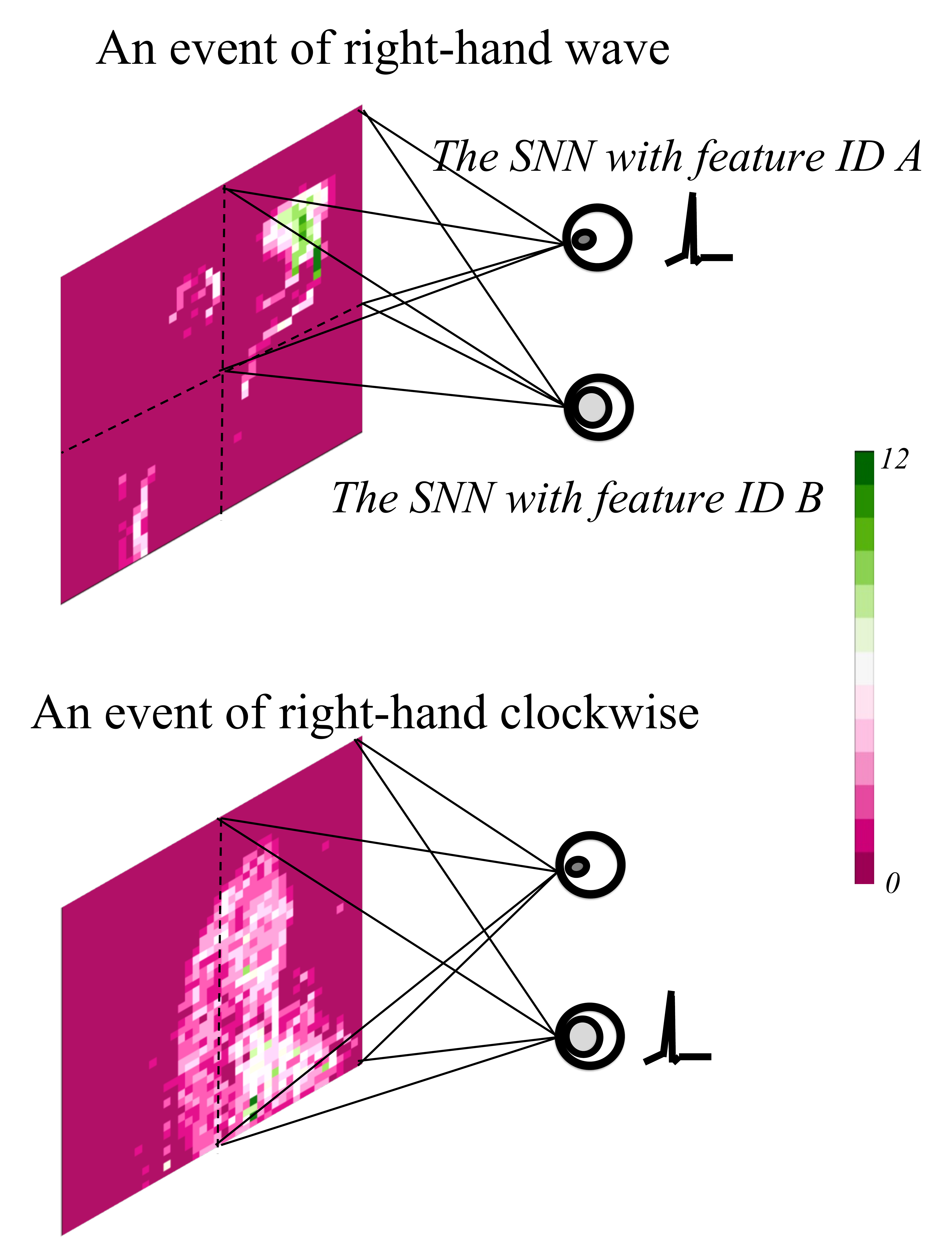} 
    \caption{The spatial SNN with feature index A and B results on events of right hand wave and right hand clockwise.}
    \label{Fig-snns-spatial_demo}
\end{figure}

Spatial SNNs computing performance is shown at Fig. \ref{Fig-snns-spatial_demo}, events of right-hand wave and right hand clock-wise are employed as examples. The results indicate that SNN with feature index A generates a spike for an event of right-hand wave, since it is only sensitive on intensive activities on a constrained area. And the SNN with feature index B generates a spike for an event of right-hand clockwise, which results from the interests of mild activities on a large area.

\begin{figure}[th]
    \centering
	\includegraphics[width=0.45\columnwidth]{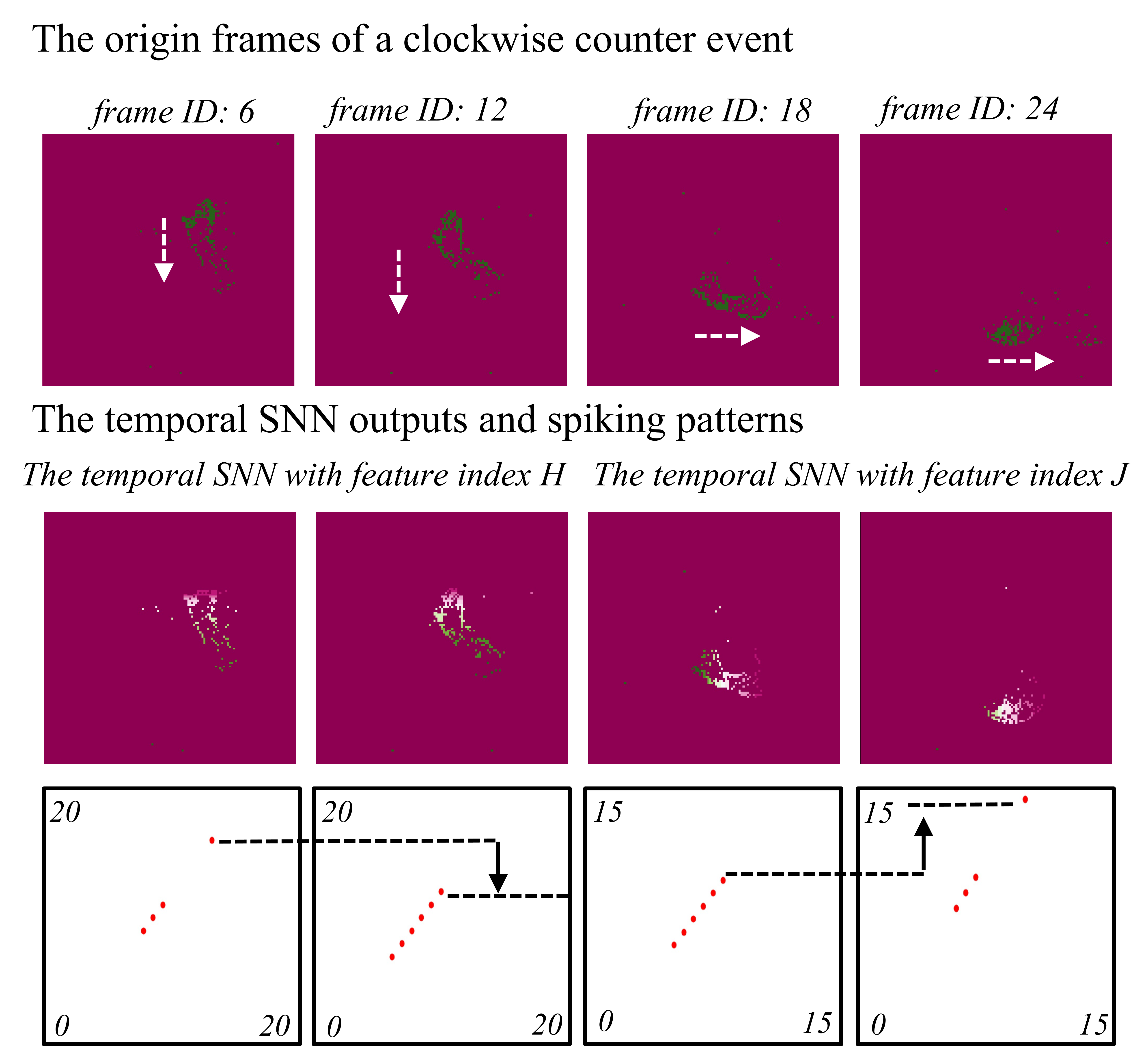} 
    \caption{The temporal SNN with feature index E performances.}
    \label{Fig-snns-temporal_demo}
\end{figure}

Temporal SNNs computing performance is shown at Fig. \ref{Fig-snns-temporal_demo}. Here we employed an event of clockwise counter as an example. Four original frame information is shown at Fig. \ref{Fig-snns-temporal_demo} top. Corresponding temporal SNNs with feature index H (top-down) and J (left-right) outputs are shown at Fig. \ref{Fig-snns-temporal_demo} middle. It is clearly seen that there is a top-down movement pattern followed by a left-right movement pattern, which are identical to the pattern sequence of clockwise counter event.

\subsection{Few-shot learning performance}

\begin{figure}[t]
	\centering
	\includegraphics[width=0.5\columnwidth]{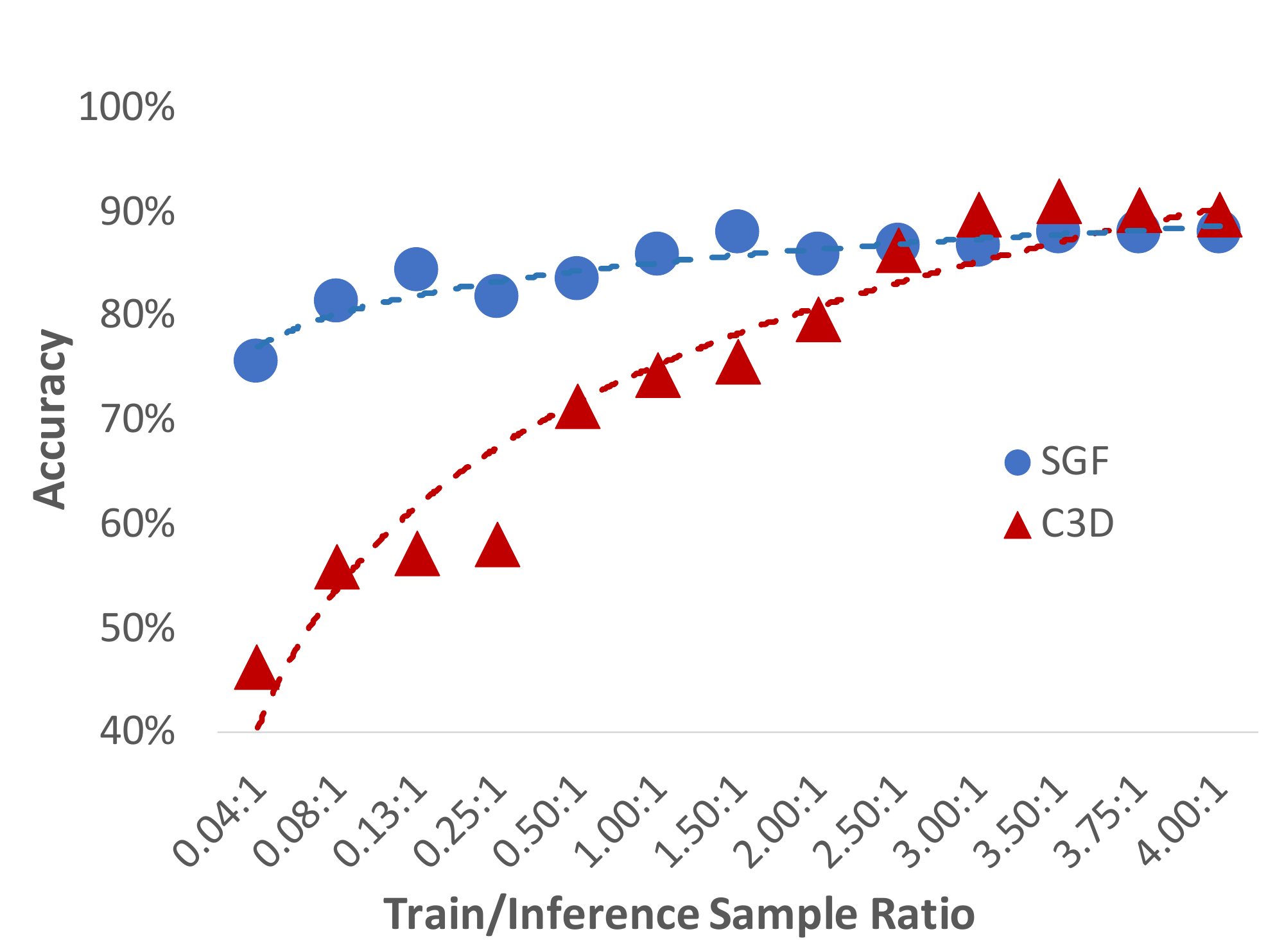} 
	\caption{A comparison of the few-shot learning performances on both SGF and C3D networks.}
	\label{Fig-few_sample}
\end{figure}

We also investigate the system's few-shot learning performance. By varying the data ratio between the training and inference stage, results are shown at Fig. \ref{Fig-few_sample}. Compared to a typical action recognition deep learning model C3D \cite{DBLP:conf/iccv/TranBFTP15} (red line), the developed SGF illustrates excellent few-shot learning performances. At a training/inference data number ratio 1.5:1 condition, the SGF  reached the highest 87.5\% accuracy, while the C3D network only has 70\% accuracy. However, there is a cross-point at a training/inference data number ratio 3.8:1. The C3D network has reached above 90\% accuracy and surpassed the SGF network. In summary, we conclude the key design principles of the developed few-shot learning paradigm: 1) a  hierarchical  structure-based network design involves with human prior knowledge; 2) SNNs for global dynamic feature detection. 

\begin{figure}[t]
	\centering
	\includegraphics[width=0.5\columnwidth]{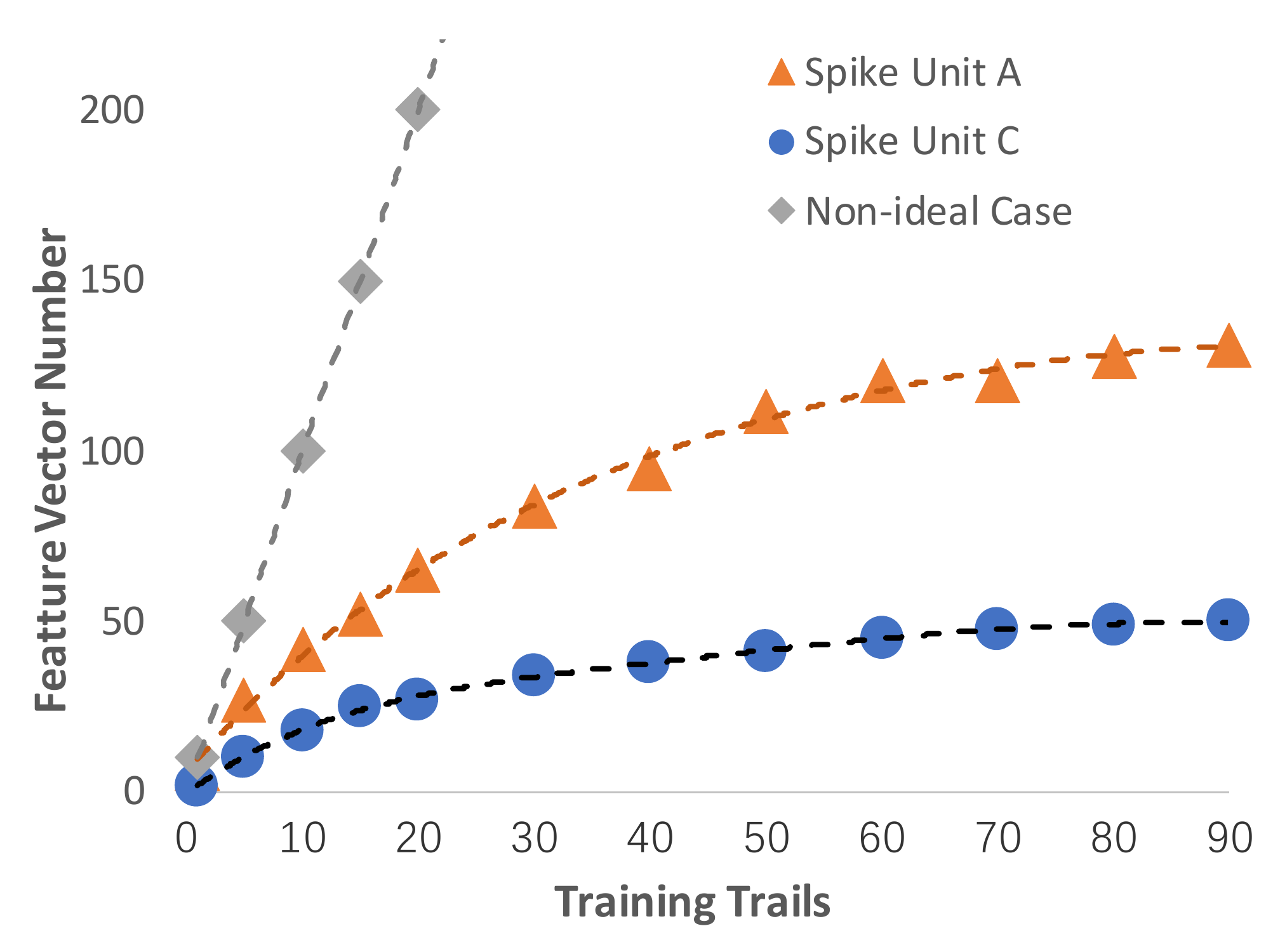} 
	\caption{The SNNs design qualities performances. the SGF unit A and C are shown in orange and blue colour, while an non-ideal design case is shown in gray colour.}
	\label{Fig-knowledge_num}
\end{figure}

\subsection{SNNs design quality}

SGF unit design qualities can be evaluated by counting generated feature vector types per training trail. Ideally, the number of feature vector types should be convergent as training trail increases. This indicates that the developed SNNs have strong feature generalization capabilities. As Fig. \ref{Fig-knowledge_num} illustrates, the SGF unit A and C feature vector numbers are convergent gradually as we expected, while a non-ideal design case is also displayed to describe the divergence issues.

\subsection{Hardware implementations}

\begin{figure}[th]
    \centering
	\includegraphics[width=0.5\columnwidth]{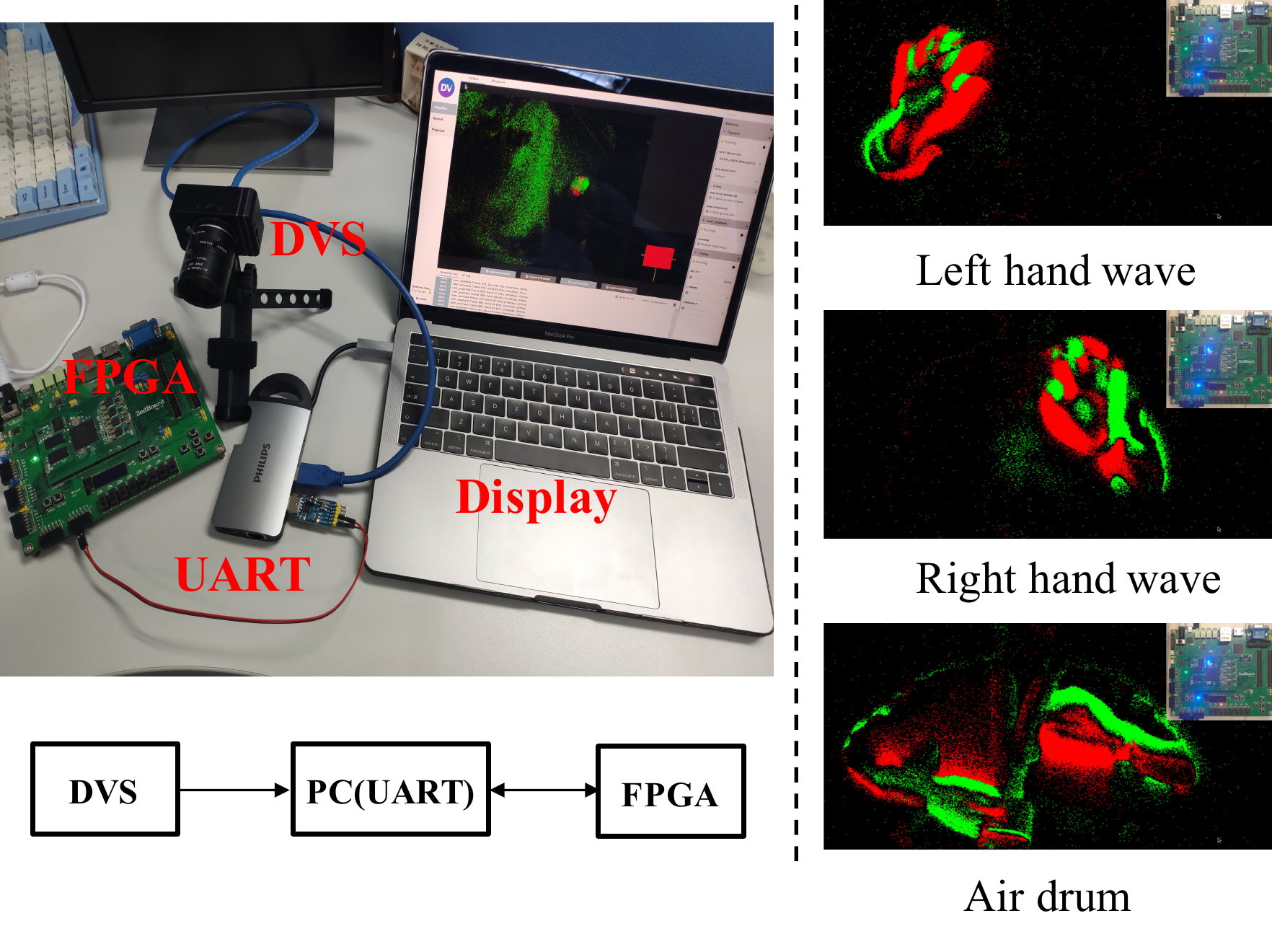} 
    \caption{The experimental setup: an SGF inference model is implemented on an FPGA ZedBoard XC7Z020-CLG484-1 for testing system hardware performances.}
    \label{fpga_demo}
\end{figure}

An SGF inference model is implemented on an FPGA ZedBoard XC7Z020-CLG484-1 for testing system hardware performance. To proof of a concept, the implemented SGF has capabilities of classifying five events. The developed hardware architecture is largely based on \cite{7293235} with modifications. The system configuration is shown at Fig. \ref{fpga_demo}(a), a DVS camera DAVIS346 (Inivation) is directed connected to a laptop HP Pro Book 430 G6 via a custom designed UART communication protocol. Three event types classification results are shown at Fig. \ref{fpga_demo} (b): left wave, right wave and air drum. The power consumption is 9mW and memory size is 99KB, which servers as a ideal candidate for edge/end-device applications.The detailed hardware implementations are shown at Appendix B.

\section{Discussion}
\label{sec:conclusion}
This work presents a novel system titled SGF which has a strong dispersity with the current mainstream deep learning networks. First, we employed a standard DVS gesture classification as a proof of concept. Regarding the training performances, the developed network can achieve the same level of accuracy with the DL under a condition of the training/inference data ratio 1.5:1. More importantly, only one training epoch is required during the learning periods, which significantly reduced the system training cost. Also, in terms of the model complexity, the SGF model size is approximately 379 times smaller than a standard CNN network, and 2.8 times smaller than an SNN. Then we implemented a developed SGF inference model on a tailor designed hardware results of limited power consummations (9mW) and memory resources (99KB). At last we draw conclusions of essential factors achieving few-shot learning paradigms: 1) a hierarchical architecture design encoded with human prior knowledge; 2) SNNs for global feature detections. At last, although the developed system capability has a considerable distance compared to the current DL network, the system shows the essential biological intelligence (e.g. few-shot learning, energy efficient, explainable) at a particular scenario. This may inspire us to design the next generation deep learning algorithms, and also raise a wider discussion among groups of computer hardware architecture, neuroscience and algorithms.

One of the major future work is that the network does not have strong generalization capabilities, which may not be suitable for processing a large-scale dataset \cite{Soomro2012UCF101AD} (e.g. UCF101). This is due to the network architecture being tailor designed for the gesture recognition task. Such a design principle of fusing human prior knowledge with SNNs global feature detection introduces a biological intelligence-based system (e.g. few-shot learning, energy efficient and explainable) but with limited flexibility. In the future, we will focus on solving the  generalization issue in various technology paths: 1) designing an SNN based Network Architecture Search (NAS) mechanism which similar to the Auto ML \cite{AutoML}; 2) introducing a reinforcement learning based agent to generate learning rules that equal to the human prior knowledge \cite{ReinforcementLearning}; 3) utilizing biological brain assembling theories to build a learning logic based architecture \cite{Papadimitriou14464, wu2022brain}.
\bibliographystyle{Frontiers-Vancouver} 
\bibliography{references}

\begin{thebibliography}{43}
\expandafter\ifx\csname natexlab\endcsname\relax\def\natexlab#1{#1}\fi
\expandafter\ifx\csname urlstyle\endcsname\relax
  \expandafter\ifx\csname doi\endcsname\relax
  \def\doi#1{doi:\discretionary{}{}{}#1}\fi \else
  \expandafter\ifx\csname doi\endcsname\relax
  \def\doi{doi:\discretionary{}{}{}\begingroup \urlstyle{rm}\Url}\fi \fi
\expandafter\ifx\csname selectlanguage\endcsname\relax
  \def\selectlanguage#1{}\fi

\bibitem[{Hu et~al.(2015)Hu, Yang, Yi, Kittler, Christmas, Li
  et~al.}]{DBLP:conf/iccvw/HuYYKCLH15}
Hu G, Yang Y, Yi D, Kittler J, Christmas WJ, Li SZ, et~al.
\newblock When face recognition meets with deep learning: An evaluation of
  convolutional neural networks for face recognition.
\newblock {\em 2015 {IEEE} International Conference on Computer Vision
  Workshop, {ICCV} Workshops 2015, Santiago, Chile, December 7-13, 2015\/}
  ({IEEE} Computer Society) (2015), 384--392.
\newblock \doi{10.1109/ICCVW.2015.58}.

\bibitem[{Krizhevsky et~al.(2012)Krizhevsky, Sutskever, and
  Hinton}]{DBLP:conf/nips/KrizhevskySH12}
Krizhevsky A, Sutskever I, Hinton GE.
\newblock Imagenet classification with deep convolutional neural networks.
\newblock Bartlett PL, Pereira FCN, Burges CJC, Bottou L, Weinberger KQ,
  editors, {\em Advances in Neural Information Processing Systems 25: 26th
  Annual Conference on Neural Information Processing Systems 2012. Proceedings
  of a meeting held December 3-6, 2012, Lake Tahoe, Nevada, United States\/}
  (2012), 1106--1114.

\bibitem[{He et~al.(2016)He, Zhang, Ren, and Sun}]{DBLP:conf/cvpr/HeZRS16}
He K, Zhang X, Ren S, Sun J.
\newblock Deep residual learning for image recognition.
\newblock {\em 2016 {IEEE} Conference on Computer Vision and Pattern
  Recognition, {CVPR} 2016, Las Vegas, NV, USA, June 27-30, 2016\/} ({IEEE}
  Computer Society) (2016), 770--778.
\newblock \doi{10.1109/CVPR.2016.90}.

\bibitem[{Purves et~al.(2014)Purves, Monson, Sundararajan, and
  Wojtach}]{Purves4750}
Purves D, Monson BB, Sundararajan J, Wojtach WT.
\newblock How biological vision succeeds in the physical world.
\newblock {\em Proceedings of the National Academy of Sciences\/} {\bf 111}
  (2014) 4750--4755.
\newblock \doi{10.1073/pnas.1311309111}.

\bibitem[{Liu et~al.(2015)Liu, Wang, Foroosh, Tappen, and
  Pensky}]{DBLP:conf/cvpr/LiuWFTP15}
Liu B, Wang M, Foroosh H, Tappen MF, Pensky M.
\newblock Sparse convolutional neural networks.
\newblock {\em {IEEE} Conference on Computer Vision and Pattern Recognition,
  {CVPR} 2015, Boston, MA, USA, June 7-12, 2015\/} ({IEEE} Computer Society)
  (2015), 806--814.
\newblock \doi{10.1109/CVPR.2015.7298681}.

\bibitem[{Wen et~al.(2016)Wen, Wu, Wang, Chen, and
  Li}]{DBLP:conf/nips/WenWWCL16}
Wen W, Wu C, Wang Y, Chen Y, Li H.
\newblock Learning structured sparsity in deep neural networks.
\newblock Lee DD, Sugiyama M, von Luxburg U, Guyon I, Garnett R, editors, {\em
  Advances in Neural Information Processing Systems 29: Annual Conference on
  Neural Information Processing Systems 2016, December 5-10, 2016, Barcelona,
  Spain\/} (2016), 2074--2082.

\bibitem[{Liu et~al.(2021)Liu, Zhao, Wang, Zou, Zhang, and
  Shi}]{DBLP:conf/glvlsi/LiuZWZZS21}
Liu S, Zhao Z, Wang Y, Zou Q, Zhang Y, Shi CR.
\newblock Systolic-array deep-learning acceleration exploring pattern-indexed
  coordinate-assisted sparsity for real-time on-device speech processing.
\newblock Chen Y, Zhirnov VV, Sasan A, Savidis I, editors, {\em {GLSVLSI} '21:
  Great Lakes Symposium on {VLSI} 2021, Virtual Event, USA, June 22-25, 2021\/}
  ({ACM}) (2021), 353--358.
\newblock \doi{10.1145/3453688.3461530}.

\bibitem[{Wang et~al.(2020)Wang, Yao, Kwok, and Ni}]{fsl_survey}
Wang Y, Yao Q, Kwok JT, Ni LM.
\newblock Generalizing from a few examples: A survey on few-shot learning.
\newblock {\em ACM Comput. Surv.\/} {\bf 53} (2020).
\newblock \doi{10.1145/3386252}.

\bibitem[{Chen et~al.(2022)Chen, Li, Wei, Zhou, and Zeng}]{HGNNFSL}
Chen C, Li K, Wei W, Zhou JT, Zeng Z.
\newblock Hierarchical graph neural networks for few-shot learning.
\newblock {\em IEEE Transactions on Circuits and Systems for Video
  Technology\/} {\bf 32} (2022) 240--252.
\newblock \doi{10.1109/TCSVT.2021.3058098}.

\bibitem[{Sung et~al.(2018)Sung, Yang, Zhang, Xiang, Torr, and
  Hospedales}]{RNFSL}
Sung F, Yang Y, Zhang L, Xiang T, Torr PH, Hospedales TM.
\newblock Learning to compare: Relation network for few-shot learning.
\newblock {\em Proceedings of the IEEE conference on computer vision and
  pattern recognition\/} (2018), 1199--1208.

\bibitem[{Lobo et~al.(2020)Lobo, Ser, Bifet, and
  Kasabov}]{DBLP:journals/nn/LoboSBK20}
Lobo JL, Ser JD, Bifet A, Kasabov NK.
\newblock Spiking neural networks and online learning: An overview and
  perspectives.
\newblock {\em Neural Networks\/} {\bf 121} (2020) 88--100.
\newblock \doi{10.1016/j.neunet.2019.09.004}.

\bibitem[{Furber and Temple(2008)}]{DBLP:series/sci/FurberT08}
Furber SB, Temple S.
\newblock Neural systems engineering.
\newblock Fulcher J, Jain LC, editors, {\em Computational Intelligence: {A}
  Compendium\/} (Springer), {\em Studies in Computational Intelligence\/}, vol.
  115 (2008), 763--796.
\newblock \doi{10.1007/978-3-540-78293-3\_18}.

\bibitem[{Wu et~al.(2018)Wu, Deng, Li, Zhu, and Shi}]{10.3389/fnins.2018.00331}
Wu Y, Deng L, Li G, Zhu J, Shi L.
\newblock Spatio-temporal backpropagation for training high-performance spiking
  neural networks.
\newblock {\em Frontiers in Neuroscience\/} {\bf 12} (2018) 331.
\newblock \doi{10.3389/fnins.2018.00331}.

\bibitem[{Amir et~al.(2017)Amir, Taba, Berg, Melano, McKinstry, di~Nolfo
  et~al.}]{DBLP:conf/cvpr/AmirTBMMNNAGMKD17}
Amir A, Taba B, Berg DJ, Melano T, McKinstry JL, di~Nolfo C, et~al.
\newblock A low power, fully event-based gesture recognition system.
\newblock {\em 2017 {IEEE} Conference on Computer Vision and Pattern
  Recognition, {CVPR} 2017, Honolulu, HI, USA, July 21-26, 2017\/} ({IEEE}
  Computer Society) (2017), 7388--7397.
\newblock \doi{10.1109/CVPR.2017.781}.

\bibitem[{Lee et~al.(2016)Lee, Delbruck, and
  Pfeiffer}]{10.3389/fnins.2016.00508}
Lee JH, Delbruck T, Pfeiffer M.
\newblock Training deep spiking neural networks using backpropagation.
\newblock {\em Frontiers in Neuroscience\/} {\bf 10} (2016) 508.
\newblock \doi{10.3389/fnins.2016.00508}.

\bibitem[{Zhang and Li(2019)}]{DBLP:conf/nips/ZhangL19}
Zhang W, Li P.
\newblock Spike-train level backpropagation for training deep recurrent spiking
  neural networks.
\newblock Wallach HM, Larochelle H, Beygelzimer A, d'Alch{\'{e}}{-}Buc F, Fox
  EB, Garnett R, editors, {\em Advances in Neural Information Processing
  Systems 32: Annual Conference on Neural Information Processing Systems 2019,
  NeurIPS 2019, December 8-14, 2019, Vancouver, BC, Canada\/} (2019),
  7800--7811.

\bibitem[{Caporale and Dan(2008)}]{STDP}
Caporale N, Dan Y.
\newblock Spike timing–dependent plasticity: A hebbian learning rule.
\newblock {\em Annual Review of Neuroscience\/} {\bf 31} (2008) 25--46.
\newblock \doi{10.1146/annurev.neuro.31.060407.125639}.
\newblock PMID: 18275283.

\bibitem[{Eliasmith(2005)}]{10.1162/0899766053630332}
Eliasmith C.
\newblock {A Unified Approach to Building and Controlling Spiking Attractor
  Networks}.
\newblock {\em Neural Computation\/} {\bf 17} (2005) 1276--1314.
\newblock \doi{10.1162/0899766053630332}.

\bibitem[{Bekolay et~al.(2014)Bekolay, Bergstra, Hunsberger, DeWolf, Stewart,
  Rasmussen et~al.}]{10.3389/fninf.2013.00048}
Bekolay T, Bergstra J, Hunsberger E, DeWolf T, Stewart T, Rasmussen D, et~al.
\newblock Nengo: a python tool for building large-scale functional brain
  models.
\newblock {\em Frontiers in Neuroinformatics\/} {\bf 7} (2014) 48.
\newblock \doi{10.3389/fninf.2013.00048}.

\bibitem[{Voelker et~al.(2019)Voelker, Kajic, and
  Eliasmith}]{DBLP:conf/nips/VoelkerKE19}
Voelker A, Kajic I, Eliasmith C.
\newblock Legendre memory units: Continuous-time representation in recurrent
  neural networks.
\newblock Wallach HM, Larochelle H, Beygelzimer A, d'Alch{\'{e}}{-}Buc F, Fox
  EB, Garnett R, editors, {\em Advances in Neural Information Processing
  Systems 32: Annual Conference on Neural Information Processing Systems 2019,
  NeurIPS 2019, December 8-14, 2019, Vancouver, BC, Canada\/} (2019),
  15544--15553.

\bibitem[{Chilkuri et~al.(2021)Chilkuri, Hunsberger, Voelker, Malik, and
  Eliasmith}]{DBLP:journals/corr/abs-2110-02402}
Chilkuri N, Hunsberger E, Voelker A, Malik G, Eliasmith C.
\newblock Language modeling using lmus: 10x better data efficiency or improved
  scaling compared to transformers.
\newblock {\em CoRR\/} {\bf abs/2110.02402} (2021).

\bibitem[{Luo and Chen(2020)}]{DBLP:journals/corr/abs-2001-10159}
Luo J, Chen J.
\newblock An internal clock based space-time neural network for motion speed
  recognition.
\newblock {\em CoRR\/} {\bf abs/2001.10159} (2020).

\bibitem[{Sussillo and Abbott(2009)}]{SUSSILLO2009544}
Sussillo D, Abbott L.
\newblock Generating coherent patterns of activity from chaotic neural
  networks.
\newblock {\em Neuron\/} {\bf 63} (2009) 544--557.
\newblock \doi{https://doi.org/10.1016/j.neuron.2009.07.018}.

\bibitem[{Imam and Cleland(2019)}]{DBLP:journals/corr/abs-1906-07067}
Imam N, Cleland TA.
\newblock Rapid online learning and robust recall in a neuromorphic olfactory
  circuit.
\newblock {\em CoRR\/} {\bf abs/1906.07067} (2019).

\bibitem[{Wu et~al.(2022)Wu, Zhao, Zhu, Chen, Xu, Li et~al.}]{wu2022brain}
Wu Y, Zhao R, Zhu J, Chen F, Xu M, Li G, et~al.
\newblock Brain-inspired global-local learning incorporated with neuromorphic
  computing.
\newblock {\em Nature Communications\/} {\bf 13} (2022) 1--14.

\bibitem[{Posch et~al.(2011)Posch, Matolin, and
  Wohlgenannt}]{DBLP:journals/jssc/PoschMW11}
Posch C, Matolin D, Wohlgenannt R.
\newblock A {QVGA} 143 db dynamic range frame-free {PWM} image sensor with
  lossless pixel-level video compression and time-domain {CDS}.
\newblock {\em {IEEE} J. Solid State Circuits\/} {\bf 46} (2011) 259--275.
\newblock \doi{10.1109/JSSC.2010.2085952}.

\bibitem[{Paulin(2004)}]{NEF}
Paulin MG.
\newblock Neural engineering: Computation, representation and dynamics in
  neurobiological systems: Chris eliasmith, charles anderson; {MIT} press
  (december 2003), {ISBN:} 0262050714.
\newblock {\em Neural Networks\/} {\bf 17} (2004) 461--463.
\newblock \doi{10.1016/j.neunet.2004.01.002}.

\bibitem[{Papadimitriou et~al.(2020)Papadimitriou, Vempala, Mitropolsky,
  Collins, and Maass}]{Papadimitriou14464}
Papadimitriou CH, Vempala SS, Mitropolsky D, Collins M, Maass W.
\newblock Brain computation by assemblies of neurons.
\newblock {\em Proceedings of the National Academy of Sciences\/} {\bf 117}
  (2020) 14464--14472.
\newblock \doi{10.1073/pnas.2001893117}.

\bibitem[{M{\"u}ller et~al.(2020)M{\"u}ller, Papadimitriou, Maass, and
  Legenstein}]{mullerENEURO.0533-19.2020}
M{\"u}ller MG, Papadimitriou CH, Maass W, Legenstein R.
\newblock A model for structured information representation in neural networks
  of the brain.
\newblock {\em eNeuro\/} {\bf 7} (2020).
\newblock \doi{10.1523/ENEURO.0533-19.2020}.

\bibitem[{Gallego et~al.(2020)Gallego, Delbruck, Orchard, Bartolozzi, Taba,
  Censi et~al.}]{9138762}
Gallego G, Delbruck T, Orchard GM, Bartolozzi C, Taba B, Censi A, et~al.
\newblock Event-based vision: A survey.
\newblock {\em IEEE Transactions on Pattern Analysis and Machine
  Intelligence\/}  (2020) 1--1.
\newblock \doi{10.1109/TPAMI.2020.3008413}.

\bibitem[{Hu et~al.(2016)Hu, Liu, Pfeiffer, and
  Delbruck}]{10.3389/fnins.2016.00405}
Hu Y, Liu H, Pfeiffer M, Delbruck T.
\newblock Dvs benchmark datasets for object tracking, action recognition, and
  object recognition.
\newblock {\em Frontiers in Neuroscience\/} {\bf 10} (2016) 405.
\newblock \doi{10.3389/fnins.2016.00405}.

\bibitem[{Rebecq et~al.(2019)Rebecq, Ranftl, Koltun, and
  Scaramuzza}]{DBLP:conf/cvpr/RebecqRKS19}
Rebecq H, Ranftl R, Koltun V, Scaramuzza D.
\newblock Events-to-video: Bringing modern computer vision to event cameras.
\newblock {\em {IEEE} Conference on Computer Vision and Pattern Recognition,
  {CVPR} 2019, Long Beach, CA, USA, June 16-20, 2019\/} (Computer Vision
  Foundation / {IEEE}) (2019), 3857--3866.
\newblock \doi{10.1109/CVPR.2019.00398}.

\bibitem[{George et~al.(2020)George, Banerjee, Dey, Mukherjee, and
  Balamurali}]{DBLP:conf/ijcnn/GeorgeBDMB20}
George AM, Banerjee D, Dey S, Mukherjee A, Balamurali P.
\newblock A reservoir-based convolutional spiking neural network for gesture
  recognition from {DVS} input.
\newblock {\em 2020 International Joint Conference on Neural Networks, {IJCNN}
  2020, Glasgow, United Kingdom, July 19-24, 2020\/} ({IEEE}) (2020), 1--9.
\newblock \doi{10.1109/IJCNN48605.2020.9206681}.

\bibitem[{Perez-Nieves et~al.(2021)Perez-Nieves, Leung, Dragotti, and
  Goodman}]{Perez-Nieves2020.12.18.423468}
Perez-Nieves N, Leung VCH, Dragotti PL, Goodman DFM.
\newblock Neural heterogeneity promotes robust learning.
\newblock {\em Nature Communications\/}  (2021).
\newblock \doi{10.1038/s41467-021-26022-3}.

\bibitem[{Xing et~al.(2020)Xing, Caterina, and Soraghan}]{scrnn}
Xing Y, Caterina GD, Soraghan J.
\newblock A new spiking convolutional recurrent neural network (scrnn) with
  applications to event-based hand gesture recognition.
\newblock {\em Frontiers in Neuroscience\/} {\bf 14} (2020) 590164.

\bibitem[{Shrestha and Orchard(2018)}]{slayer}
Shrestha SB, Orchard G.
\newblock Slayer: Spike layer error reassignment in time  (2018).

\bibitem[{Kugele et~al.(2020)Kugele, Pfeil, Pfeiffer, and Chicca}]{ann2snn}
Kugele A, Pfeil T, Pfeiffer M, Chicca E.
\newblock Efficient processing of spatio-temporal data streams with spiking
  neural networks.
\newblock {\em Frontiers in Neuroscience\/} {\bf 14} (2020) 439.

\bibitem[{Qi et~al.(2017)Qi, Su, Mo, and Guibas}]{DBLP:conf/cvpr/QiSMG17}
Qi CR, Su H, Mo K, Guibas LJ.
\newblock Pointnet: Deep learning on point sets for 3d classification and
  segmentation.
\newblock {\em 2017 {IEEE} Conference on Computer Vision and Pattern
  Recognition, {CVPR} 2017, Honolulu, HI, USA, July 21-26, 2017\/} ({IEEE}
  Computer Society) (2017), 77--85.
\newblock \doi{10.1109/CVPR.2017.16}.

\bibitem[{Tran et~al.(2015)Tran, Bourdev, Fergus, Torresani, and
  Paluri}]{DBLP:conf/iccv/TranBFTP15}
Tran D, Bourdev LD, Fergus R, Torresani L, Paluri M.
\newblock Learning spatiotemporal features with 3d convolutional networks.
\newblock {\em 2015 {IEEE} International Conference on Computer Vision, {ICCV}
  2015, Santiago, Chile, December 7-13, 2015\/} ({IEEE} Computer Society)
  (2015), 4489--4497.
\newblock \doi{10.1109/ICCV.2015.510}.

\bibitem[{Luo et~al.(2016)Luo, Coapes, Mak, Yamazaki, Tin, and
  Degenaar}]{7293235}
Luo J, Coapes G, Mak T, Yamazaki T, Tin C, Degenaar P.
\newblock Real-time simulation of passage-of-time encoding in cerebellum using
  a scalable fpga-based system.
\newblock {\em IEEE Transactions on Biomedical Circuits and Systems\/} {\bf 10}
  (2016) 742--753.
\newblock \doi{10.1109/TBCAS.2015.2460232}.

\bibitem[{Soomro et~al.(2012)Soomro, Zamir, and Shah}]{Soomro2012UCF101AD}
Soomro K, Zamir AR, Shah M.
\newblock Ucf101: A dataset of 101 human actions classes from videos in the
  wild.
\newblock {\em ArXiv\/} {\bf abs/1212.0402} (2012).

\bibitem[{He et~al.(2021)He, Zhao, and Chu}]{AutoML}
He X, Zhao K, Chu X.
\newblock Automl: A survey of the state-of-the-art.
\newblock {\em Knowledge-Based Systems\/} {\bf 212} (2021) 106622.
\newblock \doi{https://doi.org/10.1016/j.knosys.2020.106622}.

\bibitem[{Williams(1992)}]{ReinforcementLearning}
Williams RJ.
\newblock Simple statistical gradient-following algorithms for connectionist
  reinforcement learning.
\newblock {\em Machine Learning\/} {\bf 8} (1992) 229--256.

\end{thebibliography}

\end{document}